\documentclass[10pt,twocolumn,letterpaper]{article}

\usepackage{cvpr}              %

\usepackage{graphicx}
\usepackage{amsmath}
\usepackage{amssymb}
\usepackage{booktabs}

\usepackage{times}
\usepackage{epsfig}
\usepackage{multirow}
\usepackage{adjustbox}
\usepackage{tabu}
\usepackage{xcolor}
\usepackage{footnote, eucal}
\usepackage[linesnumbered, ruled]{algorithm2e}
\usepackage{algpseudocode}%
\usepackage{bm}
\usepackage{array}
\usepackage{enumitem}
\usepackage{verbatim}
\usepackage{caption}
\usepackage{algorithmicx}
\usepackage{comment}
\usepackage{lipsum}
\usepackage{stringstrings}
\usepackage{makecell}
\usepackage{url}
\usepackage[accsupp]{axessibility}  %

\usepackage[pagebackref,breaklinks,colorlinks]{hyperref}

\usepackage[capitalize]{cleveref}
\crefname{section}{Sec.}{Secs.}
\Crefname{section}{Section}{Sections}
\Crefname{table}{Table}{Tables}
\crefname{table}{Tab.}{Tabs.}

\def\cvprPaperID{4040} %
\def\confName{CVPR}
\def\confYear{2022}

\def\eg{\textit{e.g.}}
\def\ie{\textit{i.e.}}

\def\vs{\textit{vs.\ }}
\def\aka{\textit{aka.\ }}

\def\OurMethodTwo{{LP}\xspace}
\def\OurMethodThree{{PLI}\xspace}

\def\OurWholeMethod{{ISE}\xspace}
\def\market{{Market-1501}\xspace}

\def\msmt{{MSMT17}\xspace}
\newcommand{\myparagraph}[1]{{\setlength{\parskip}{0.3em} \noindent \textbf {#1}}}

\begin{document}
\title{Implicit Sample Extension for Unsupervised Person Re-Identification}

\author{Xinyu Zhang\textsuperscript{\rm 1\thanks{Equal contribution. 
}}, Dongdong Li\textsuperscript{\rm 3,1$*$}, Zhigang Wang\textsuperscript{\rm 1}, Jian Wang\textsuperscript{\rm 1},  Errui Ding\textsuperscript{\rm 1}, \\ Javen Qinfeng Shi\textsuperscript{\rm 2}, Zhaoxiang Zhang\textsuperscript{\rm 3,4}, Jingdong Wang\textsuperscript{\rm 1\thanks{
Corresponding author. 
}}\\
\textsuperscript{\rm 1}Baidu VIS, China  \textsuperscript{\rm 2}The University of Adelaide, Australia \textsuperscript{\rm 3}Institute of Automation, CAS \& UCAS, China \\
\textsuperscript{\rm 4} Centre for Artificial Intelligence and Robotics, HKISI\_CAS, China\\
{\tt\small \{zhangxinyu14,wangzhigang05,wangjian33,dingerrui,wangjingdong\}@baidu.com
}\\
{\tt\small \{lidongdong2019,zhaoxiang.zhang\}@ia.ac.cn, \tt\small javen.shi@adelaide.edu.au
}
}
\maketitle

\begin{abstract}
Most existing unsupervised person re-identification (Re-ID) methods use clustering to generate pseudo labels for model training. Unfortunately, clustering sometimes mixes different true identities together or splits the same identity into two or more sub clusters. Training on these noisy clusters substantially hampers the Re-ID accuracy. 
Due to the limited samples in each identity, we suppose there may lack some underlying information to well reveal the accurate clusters. 
To discover these information, we propose an Implicit Sample Extension (\OurWholeMethod) method to generate what we call support samples around the cluster boundaries. 
Specifically, we generate support samples from actual samples and their neighbouring clusters in the embedding space through a progressive linear interpolation (PLI) strategy. PLI controls the generation with two critical factors, i.e., 1) the direction from the actual sample towards its K-nearest clusters and 2) the degree for mixing up the context information from the K-nearest clusters. 
Meanwhile, given the support samples, ISE further uses a label-preserving loss to pull them towards their corresponding actual samples, so as to compact each cluster.
Consequently, ISE reduces the ``sub and mixed'' clustering errors, thus improving the Re-ID performance. 
Extensive experiments demonstrate that the proposed method is effective and achieves state-of-the-art performance for unsupervised person Re-ID.
Code is available at: \url{https://github.com/PaddlePaddle/PaddleClas}.

\end{abstract}

\begin{figure}[t]
\centering
\includegraphics[trim =0mm 0mm 0mm 0mm, clip, width=1.0\linewidth]{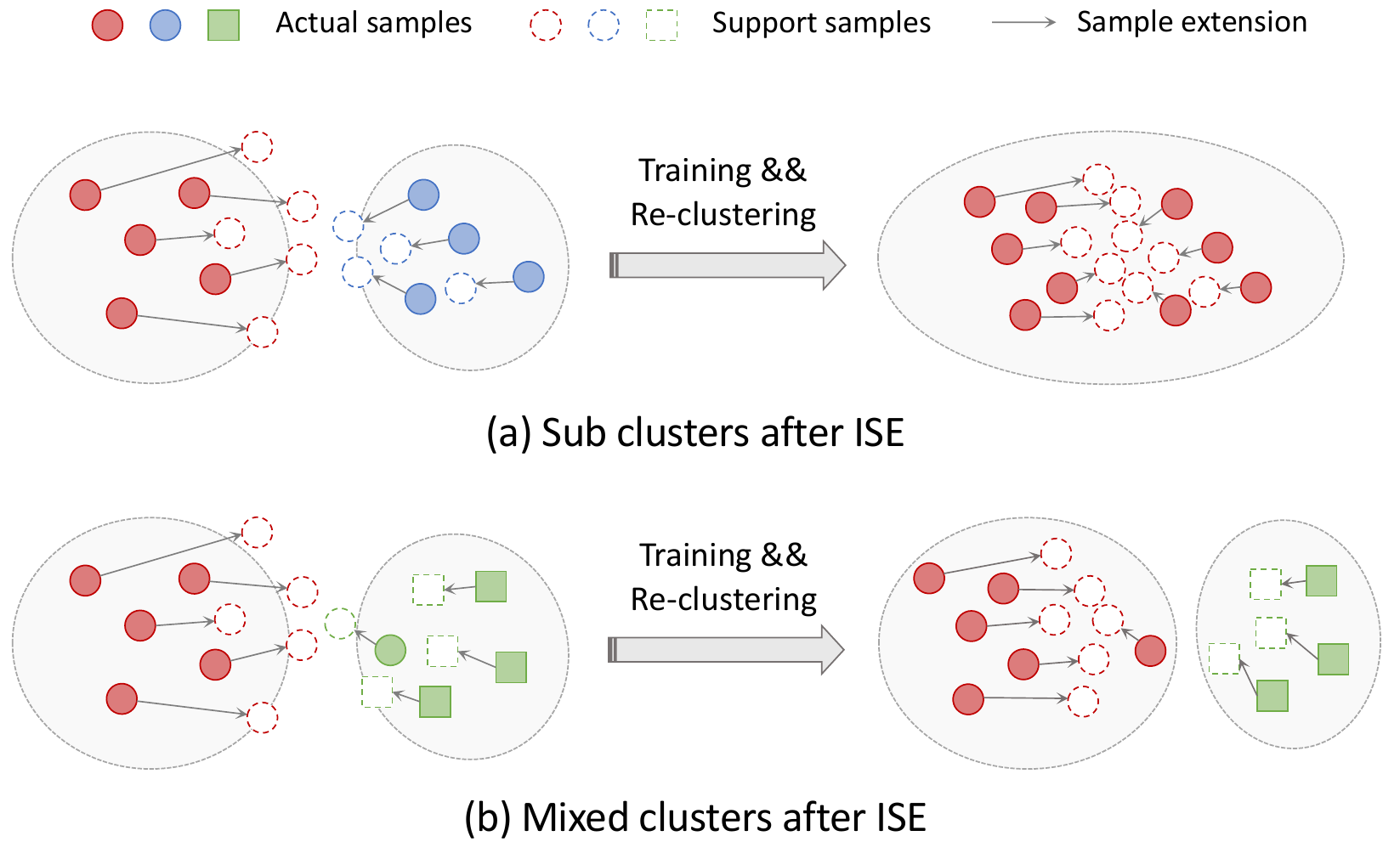}
\caption{Clustering behaviour changes with our implicit sample extension (\OurWholeMethod). Different
shapes represent different ground-truth identities, while different colors stand for different pseudo labels (best viewed in color). 
Usually, clustering may produce (a) {\bf sub clusters} that samples share the same true identity but are split apart as two or more clusters; and (b) {\bf mixed clusters} that samples with different identities 
are mixed to the same cluster. Our support samples help to merge the sub clusters as one and split the mixed clusters apart, thus improving the performance. 
}
\vspace{-0.5cm}
\label{fig:figure1}
\end{figure}

\section{Introduction}
\label{sec:intro}

Unsupervised person re-identification (Re-ID) aims to learn person appearance features without annotations. It gains increasing attention and popularity due to its wide practical applications in the real world. One kind of works \cite{udatp,fu2019self,zhang2019self,zhong2019invariance,UMSDA,dai2021idm,zhang2020memorizing} attempt to transfer knowledge from existing labeled data to unlabeled target data, known as unsupervised domain adaptation (UDA). Another kind of works rely on unsupervised learning (USL) to learn representations purely from unlabeled images, which is more data-friendly than UDA. Most USL Re-ID methods \cite{dai2021cluster,ge2020self,chen2021ice} follow an iterative two-stage training procedure: 1) using clustering \cite{macqueen1967some, ester1996density} to generate pseudo labels as the supervision for the next step;
2) training the Re-ID model using the pseudo labels.
Here we focus on investigating the USL Re-ID, and follow the two-stage pipeline.

Clustering is critical in the aforementioned USL and has attracted great attention from existing USL methods. For instance, ACT \cite{asymmetric} utilizes two models that try to refine pseudo labels for each other. SpCL \cite{ge2020self} employs a self-paced learning to gradually generate more reliable clusters for training. Despite the progresses, these methods still suffer from noisy pseudo labels, especially the \textit{sub} and \textit{mixed} clusters shown in the left part of Figure~\ref{fig:figure1}.
Based on this observation, we suppose that some underlying information may be lost in the existing data distribution due to the limited samples in each identity.
If we can compensate for the missing information, the clustering quality can be improved in the next step.
For example, if there exists intermediate variations among two sub clusters with the same identity, they can be correctly merged into one cluster.

To achieve this goal, we propose an implicit sample extension (\OurWholeMethod) method, progressively synthesizing supplementary samples to improve the context representation for each cluster.
We name these supplementary samples as \textit{support samples}, since they often patrol around the cluster boundaries.
Inspired by \cite{upchurch2017deep,bengio2013better,wang2021regularizing} that deep features are usually linearized, we propose a progressive linear interpolation operation (\OurMethodThree) to guide the generation of support samples with two factors, \ie, \textit{direction} and \textit{degree}. 
Specifically, 
the direction factor controls that the generation of support samples is from a actual sample to its $K$-nearest neighbor clusters in the embedding space.
Support samples in these directions are more meaningful, since neighbor clusters are more likely to have the above clustering problems.
The degree factor decides how much context information from $K$-nearest clusters should be incorporated by support samples.
We increases the degree progressively to fill the cluster gap with farther and farther support samples. 
This strategy can prevent the training collapse caused by aggressive support samples, especially in the early training stages.
Except for PLI, we also propose a label-preserving (LP) contrastive loss to enforce support samples close to their originals to compact each cluster.
After the sample generation, we regard support samples as actual ones to let them participate in the model training.

We observe that with the help of support samples, the data distribution is refined, and spurious clustering behaviours, including the sub and mixed clusters, can be alleviated (see experiments for more details).
In addition, different from GAN (General Adversarial Network) based methods \cite{deng2018image,zou2020joint}, ISE is parameter-free and performs the sample generation implicitly in the embedding space instead of explicitly in the image pixel space.
Therefore, the support samples can be readily utilized in loss functions without any feature extraction procedure.
These merits make the proposed ISE a very efficient method.
In summary, our contributions are as follows:
\begin{itemize}
\itemsep -.051cm
\item We propose a novel implicit sample extension (ISE) method for USL person Re-ID. 
The generated support samples from ISE provide complementary information, which can nicely handle the issues of sub and mixed clustering errors;
\item We present a novel progressive linear interpolation (PLI) strategy and a label-preserving contrastive loss (LP) for the support sample generation;
\item We conduct comprehensive experiments and analyses to
show the effectiveness of ISE, and ISE outperforms the current state-of-the-arts by a large margin.
\end{itemize}

\section{Related Work}
\noindent\textbf{Unsupervised Person Re-ID.}
Unsupervised person Re-ID methods are generally classified into two categories. 
The first is the unsupervised domain adaptation (UDA) methods, which transfers knowledge from labeled source data to unlabeled target data. These methods still need to annotate partial data.
Another one is the purely unsupervised learning (USL) Re-ID without any identity annotation that is more data-friendly. 
In this paper, we focus on the USL Re-ID.
Most recent works focus on pseudo label generation and framework design. 
BUC \cite{lin2019bottom} presents a bottom-up clustering scheme to gradually merge single sample into bigger clusters.
MMCL \cite{wang2020unsupervised} 
proposes a multi-label classification regime to construct relationships among individual images. 
SpCL \cite{ge2020self} employs self-paced learning to gradually generate more reliable clusters. Both CycAs \cite{CysAs} and TSSL \cite{TSSL} use tracklet information to predict pseudo labels more correctly. SoftSim \cite{lin2020unsupervised} directly uses similarity-based soft labels to train the Re-ID model. 
More recently, ICE \cite{chen2021ice} leverages inter-instance pairwise similarity to enhance the contrastive learning. Cluster-Contrast \cite{dai2021cluster} stores features and computes contrastive loss in the cluster level to update each cluster consistently. Different from these works, our method emphasises on the issues of sub and mixed clusters, and
aims to compensate for some missing information to refine cluster results and improve the performance.

\noindent\textbf{Image Generation.}
Image generation is a common strategy to enrich limited data. Mixup \cite{zhang2017mixup} and cutmix \cite{cutmix} directly fuses two images or image patches as data augmentations. Viewmaker \cite{viewmaker} yields diverse views by adding learned perturbations. SPGAN \cite{deng2018image} proposes a similarity-preserved GAN for style transfer between two domains. DGNet++ \cite{zou2020joint} disentangles structures and appearances of person images, and re-fuses them to reconstruct the new ones. Chen \etal \cite{chen2020joint} use 3D meshes to synthesize more views of an person image to learn view-invariant features. Different from these works, our method is feature-level, parameter-free and can be used to alleviate the inaccurate clustering results.

\begin{figure*}[t]
\centering
\includegraphics[trim =0mm 0mm 0mm 0mm, clip, width=1.0\linewidth]{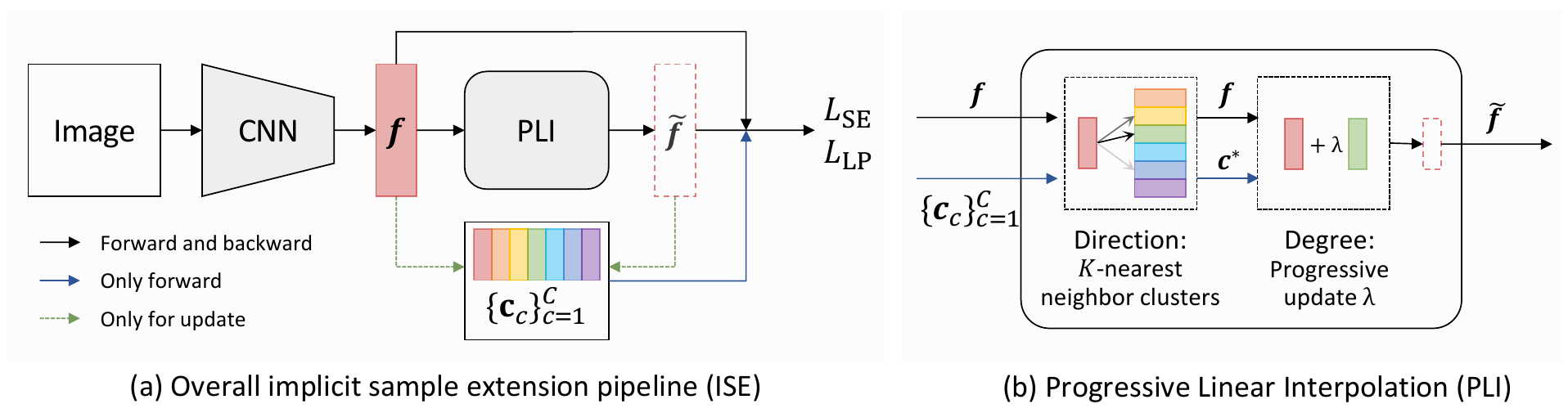}
\setlength{\abovecaptionskip}{-0.3cm} 
\setlength{\belowcaptionskip}{-0.33cm}  %
\caption{
Overview of (a) the overall implicit sample extension pipeline (\OurWholeMethod) and the details of (b) the progressive linear interpolation (\OurMethodThree) strategy.
For a specific sample feature $\pmb{f}$, 
we apply \OurMethodThree to generate the support sample $\widetilde{\pmb{f}}$, which 
are used to optimize the model with the sample extension loss $L_\mathrm{SE}$ and the label-preserving loss $L_\mathrm{LP}$.
As for the generation process,
\OurMethodThree finds the $K$-nearest cluster centroids $\pmb{c}^{*}$ as the generation \textit{direction}, and adopts the progressively updated $\lambda$ to control the generation \textit{degree}.
More details are in Algorithm~\ref{algo}.
}
\label{fig:figure2}
\end{figure*}

\section{Method}
In this section, we first set up a general formalization of existing clustering-based USL Re-ID methods, and introduce our baseline.
We then elaborate the details of our implicit sample extension (\OurWholeMethod) method, including the progressive linear interpolation strategy (\OurMethodThree) and the label-preserving scheme (\OurMethodTwo).

\subsection{Preliminary}
Let $\mathcal{X}=\{ x_1, x_2, ..., x_N\}$ denote an unlabeled Re-ID dataset, where $x_i$ is the $i$-th image and $N$ is the total image number.
In the USL Re-ID task, we aim to train a deep neural network $f_{\theta}(\cdot)$ to project an image from the data point space $\pmb{\mathcal{X}}$ into an embedding space $\pmb{\mathcal{F}}$.
Given a specific sample $x$, $f_{\theta}(x)$ is the extracted feature from the model.
Here we omit the model parameters $\theta$ and employ $\pmb{f}=f(x)$ as the feature embedding.
$\pmb{f}\in\mathbb{R}^{d}$ is a $d$-dimensional vector.

Recent clustering-based USL Re-ID methods~\cite{zhong2020learning,ge2020self,dai2021cluster,dai2021idm,wang2020camera,chen2021ice,fu2019self,wang2020unsupervised,zhang2019self} utilize K-means~\cite{macqueen1967some} or DBSCAN~\cite{ester1996density} to assign pseudo labels for unlabeled samples.
After then, we obtain a ``labeled'' dataset $\mathcal{X^{'}}=\{ (x_1, y_1), (x_2, y_2), ..., (x_{N^{'}}, y_{N^{'}})\}$, where $y_i\in\{1, \dots, C\}$ is the pseudo label of the $i$-th selected image and $C$ is the total cluster number.
Here are $N^{'}$ images with valid pseudo labels, while others are ignored.
With $\mathcal{X^{'}}$, we can optimize the model under the form of the supervised learning.
Items in the same cluster are considered as the same person, serving as positive samples.
Others in different clusters are negatives.
Based on this, existing methods~\cite{zhong2020learning,ge2020self,dai2021cluster,wang2020camera} employ the InfoNCE loss function~\cite{oord2018representation} for the model training.
Despite the diverse variants in different approaches,
we summarize the loss functions as a general formulation:
\begin{equation}
\centering
\begin{aligned}
L_{\mathrm{info}}=-\mathrm{log}\frac{\mathrm{exp}(\mathrm{sim}(\pmb{f}\cdot \pmb{m}_{+})/\tau )}{\sum_{m=1}^{M}\mathrm{exp}(\mathrm{sim}(\pmb{f}\cdot \pmb{m}_{m})/\tau )},
\end{aligned}
\label{eq:infonce_loss}
\end{equation} 
where $\pmb{m}_{m}$ is the $m$-th entry of the memory bank $\pmb{\mathrm{M}}$, and $\pmb{m}_{+}$ shares the same pseudo label with $\pmb{f}$.
$\mathrm{sim}(\pmb{u}, \pmb{v})=\pmb{u}^{\mathrm{T}}\pmb{v}/ \left|\left| \pmb{u} \right|\right| \left|\left| \pmb{v} \right|\right|$ denotes the cosine similarity between two vectors $\pmb{u}$ and $\pmb{v}$.
Some methods~\cite{zhong2020learning,ge2020self} construct the memory bank $\pmb{\mathrm{M}}\in \mathbb{R}^{d\times N}$ with the instance (\aka sample) feature.
In this way, $M$ can be the whole sample number $N$ or the mini-batch size $B$.
Other methods~\cite{dai2021cluster,wang2020camera} alternate to form the memory bank $\pmb{\mathrm{M}}\in \mathbb{R}^{d\times C}$ with the cluster centroids.
Here $M$ is equal to the cluster number $C$.

Meanwhile, these methods also design various strategies for the memory update.
In general, they adopt the momentum update strategy, which is formulated as follows:
\begin{equation}
\centering
\begin{aligned}
\pmb{m} \leftarrow \mu \cdot \pmb{m} + (1 - \mu )\cdot \pmb{f},
\end{aligned}
\label{eq:memory_update}
\end{equation} 
where $\pmb{f}$ is the encoded feature that has the same identity with the memory entry $\pmb{m}$.
Specifically, $\pmb{f}$ is utilized to update its own instance entry in \cite{zhong2020learning,ge2020self}, while to update its own cluster entry in \cite{dai2021cluster,wang2020camera}.
Cluster-Contrast~\cite{dai2021cluster} further uses the batch hardest sample for the memory update\footnote{\cite{dai2021cluster} changes to use every sample in the mini-batch for the momentum updating in the updated version. 
}. 

In this paper, we follow Cluster-Contrast~\cite{dai2021cluster} to establish our \textit{baseline}, in which the memory bank is constructed by the cluster centroids and the hardest/every sample in one mini-batch is used for the memory updating. 
The detailed training settings and the differences with Cluster-Contrast are described in Sec.~\ref{sec:implement_detail}.

\subsection{Progressive Linear Interpolation}
However, the above clustering-based USL person Re-ID methods still suffer from sub clusters and mixed clusters.
\textcolor{black}{We suppose that the underlying missing information in the embedding space causes inaccurate clusters.}
Our hypothesis lies in that the clustering quality can be improved in the next step by involving richer information as described in Sec.\ref{sec:intro}.
Based on this motivation, we present a \textit{progressive linear interpolation} (PLI) method as follows.

Given a sample feature $\pmb{f}$,
we generate its corresponding support sample $\widetilde{\pmb{f}}$ by a linear interpolation operation:
\begin{equation}
\centering
\begin{aligned}
\widetilde{\pmb{f}}=\pmb{f}+\lambda \Delta \pmb{f}, 
\end{aligned}
\label{eq:sample_extension}
\end{equation} 
where $\Delta \pmb{f}$ controls the \textit{direction} and $\lambda$ controls the \textit{degree}, which are two important factors during the sample extension.
From the perspective of problematic clusters, the ideal support samples are expected to be informative to distribute around the decision boundary.
Thus we design the following direction and degree for the sample generation.

\myparagraph{Direction.}
Neighbor clusters are more likely to have misclassified samples.
We thus aim to generate support samples distributed among two similar clusters.
Specifically,
we first find $K$-nearest neighbor clusters for $\pmb{f}$ based on the cosine similarities, which are calculated with $\pmb{f}$ and the whole memory bank $\pmb{\mathrm{M}}\in \mathbb{R}^{d\times C}$.
Then the directions of support samples are established as vectors from the cluster centroid $\pmb{c}$ of $\pmb{f}$ to the $K$-nearest ones.
Here, we set $K=1$ as an example and denote the $1$-nearest cluster centroid as $\pmb{c}^{*}$.
The direction $\Delta \pmb{f}$ is thus formulated as follows:
\begin{equation}
\centering
\begin{aligned}
\Delta \pmb{f}=\frac{1}{2}(\pmb{c}^{*}\ - \pmb{c}).
\end{aligned}
\label{eq:se_direction}
\end{equation}
In this way, the generated support samples have rich context semantics from both itself and its neighbor clusters, which are informative as the complements. The direction thus controls \textit{what} information is worth compensating for. 
For $K>1$, each neighbor cluster centroid decides a direction factor.
We generate a respective support sample for each direction, and there are total $K$ support samples for $\pmb{f}$.

\myparagraph{Degree.}
In addition, the degree is also essential in the sample extension.
The degree decides \textit{how much} information of neighbor clusters should be incorporated in support samples.
Too large $\lambda$ makes support samples far from their original clusters, misleading the model training.
Too small $\lambda$ inversely generates almost useless support samples
since they are similar to the originals. 
Therefore, we perform a \textit{progressive} update on $\lambda$ to gradually involve more context information as the training goes on.
Specifically, we set $\lambda$ as a variable of the training iteration, which increases logarithmically over the training proceeds:
\begin{equation}
\centering
\begin{aligned}
\lambda = \frac{\lambda_0}{2} \log (\frac{e-1}{T}\cdot t + 1),
\end{aligned}
\label{eq:se_degree}
\end{equation}
where $t$/$T$ represents the current/total iteration.
$\lambda_0$ is the base degree.
In the early stages, Eq.~\ref{eq:se_degree} enforces support samples near to its original sample to %
avoid introducing noise when features are not good enough.
As the training goes on, feature representations become more robust.
In that time, farther support samples, involving richer contexts from neighbor clusters, 
are necessary to include harder underlying information and augment the data distribution.

After the sample extension, support samples can be regarded as actual features for the model optimization.
We then modify the InfoNCE loss (Eq.~\ref{eq:infonce_loss}) as follows:
\begin{equation}
\centering
\begin{aligned}
L_{\mathrm{SE}}=-\mathrm{log}\frac{\mathrm{exp}(\mathrm{sim}(\widehat{\pmb{f}}\cdot \pmb{c})/\tau_{1} )}{\sum_{c=1}^{C}\mathrm{exp}(\mathrm{sim}(\widehat{\pmb{f}}\cdot \pmb{c}_{c})/\tau_{1} )},
\end{aligned}
\label{eq:infonce_se_loss}
\end{equation} 
where $\widehat{\pmb{f}}=\{\pmb{f}; \widetilde{\pmb{f}}\}$, both original samples and support samples contribute to the model optimization.
$\pmb{c}_{c}$ is the $c$-th cluster centroid. $\tau_{1}$ is a temperature hyper-parameter.
At the same time, we utilize $\widehat{\pmb{f}}$ to update the corresponding cluster centroid $\pmb{c}$ in Eq.~\ref{eq:memory_update}.
Here $\pmb{m}_c$ stores the cluster centroid $\pmb{c}_c$. In this circumstance, each cluster can tune its representations by considering more context information, other than directly fitting to the inaccurate clustering results.

\setlength{\textfloatsep}{0.15cm}
\begin{algorithm}[t]
	\begin{footnotesize}
		\SetAlgoLined
        \SetKwInOut{Input}{Input}
        \SetKwInOut{Output}{Output}
        \SetKwInput{Require}{Require}
        \Require{network $f_{\theta}$; cluster number $C$; iteration number from $T_0$ to $T_t$ in one epoch; 
        batch size $B$; cluster number in a mini-batch $C_{B}$.}
        \Input{training dataset with pseudo labels $\mathcal{X}^{'}=\{ (x_i, y_i)\}_{i=1}^{N^{'}}$, where $y_i\in\{1, \dots, C\}$;
        cluster centroids $\{\pmb{c}_c\}_{c=1}^{C}$.
		}
		\For{$t=T_0$ \KwTo $T_t$}
		{
        Sample a mini-batch $\mathcal{B}=\{(x_i, y_i)\}_{i=1}^{B}$, $y_i\in \{c_i\}_{i=1}^{C_B}$\;
		Extract embedding features $\pmb{F}=\{\pmb{f}_i\}_{i=1}^{B}$ on $\mathcal{B}$\;
		Define the generation direction by \OurMethodThree
		according to Eq.~\ref{eq:se_direction}\;
		Adjust the generation degree by \OurMethodThree according to Eq.~\ref{eq:se_degree}\;
		Generate support samples $\{\widetilde{\pmb{f}}_i\}_{i=1}^{B}$\ according to Eq.~\ref{eq:sample_extension}\;
		Calculate $L_{\mathrm{SE}}$ and $L_{\mathrm{LP}}$ according to Eq.~\ref{eq:infonce_se_loss} and Eq.~\ref{eq:lp_loss}\;
		Update cluster centroids $\{\pmb{c}_i\}_{i=1}^{C_B}$\;
		Update the network parameter $\theta$\;
        }
        \Output{updated network $f_{\theta}$ and cluster centroids $\{\pmb{c}_c\}_{c=1}^{C}$.}
		\caption{\OurWholeMethod in one epoch}\label{algo}
	\end{footnotesize}
\end{algorithm}
\setlength{\floatsep}{0.15cm}

\subsection{Label-Preserving Scheme}
\textcolor{black}{The generated support samples are helpful to extend cluster boundaries by including abundant context information.}
However, these support samples are not expected to change so much that they will become outliers of the original clusters. Except for the generation degree of support samples, we also propose a label-preserving (LP) loss to enforce them to be close to their original samples:
\begin{equation}
\centering
\begin{aligned}
L_{\mathrm{LP}}=-\mathrm{log}\frac{\mathrm{exp}(\mathrm{sim}(\pmb{f}\cdot \widetilde{\pmb{f}}^{+})/\tau_{2} )}{\sum_{c=1}^{C_B}\mathrm{exp}(\mathrm{sim}(\pmb{f}\cdot \widetilde{\pmb{f}}^{-}_{c})/\tau_{2} )},
\end{aligned}
\label{eq:lp_loss}
\end{equation} 
where $C_B$ is the cluster number of a mini-batch.
$\widetilde{\pmb{f}}^{+}$
is the hardest positive support sample for a given $\pmb{f}$, and
$\widetilde{\pmb{f}}^{-}_{c}$
is the hardest negative one in the $c$-th cluster of a mini-batch.
$\tau_{2}$ is a temperature hyper-parameter in Eq.~\ref{eq:lp_loss}.

With the proposed label-preserving loss,
support samples are constrained to be similar to their own cluster members, and dissimilar to samples from other clusters.

\subsection{Implicit Sample Extension}
We name our whole method as implicit sample extension (\OurWholeMethod), in which support samples are generated implicitly in the embedding space and dynamically changed on-the-fly.
Overall, we first generate support samples via the progressive linear interpolation method,
and then utilize them as actual samples to help the model optimization. 
Our final loss function is as follows:
\begin{equation}
\centering
\begin{aligned}
L=L_{\mathrm{SE}}+\beta L_{\mathrm{LP}},
\end{aligned}
\label{eq:overall_loss}
\end{equation} 
where $\beta$ is the loss weight, balancing two loss functions $L_{\mathrm{SE}}$ and $L_{\mathrm{LP}}$.
The details of \OurWholeMethod are in Algorithm~\ref{algo}.

\section{Experiments}

\subsection{Datasets and Evaluation Protocol}
\myparagraph{Datasets.} We evaluate our proposed method on \market~\cite{zheng2015scalable}
and \msmt~\cite{wei2018person}.
\market includes 32,668 images of 1,501 identities with 6 camera views. 
There are 12,936 images of 751 identities for training and 19,732 images of 750 identities for test.
\msmt contains 126,441 images from 4,101 identities captured by 15 cameras. The training set has 32,621 images of 1,041 identities and the test set has 93,820 images of 3,060 identities. 

\myparagraph{Evaluation protocol.} 
We use the cumulative match characteristic (CMC) curve~\cite{gray2007evaluating} and the mean average precision (mAP)~\cite{bai2017scalable} as the evaluation metrics.
There is no post-processing operations in testing, \eg, reranking~\cite{zhong2017re}.

\subsection{Implementation Details}\label{sec:implement_detail}
The backbone of the Re-ID model is ResNet-50~\cite{he2016deep} pretrained on ImageNet~\cite{deng2009imagenet}.
The model modification follows Cluster-Contrast~\cite{dai2021cluster}.
The size of the input image is 256$\times$128 and the batch size is 256.
At the beginning of each epoch, we perform the DBSCAN clustering to generate pseudo labels.
The maximum distance between two samples in DBSCAN is set to 0.4 in \market and 0.7 in \msmt.
We initial the memory bank $\pmb{\mathrm{M}}$ with cluster centroids and re-initialize it at the start of each epoch.
The memory update rate $\mu$ is empirically set to 0.2 in \market and 0.1 in \msmt.
The loss weight $\beta$ is set to 0.1.
The temperature factors $\tau_1$ and $\tau_2$ are 0.05 and 0.6, respectively.
We adopt Adam~\cite{kingma2014adam} optimizer, and the weight decay is 5e-4.
The initial learning rate is 3.5e-4, and the base degree $\lambda_0$ is 1.0.
For \market, the total epoch is 70, and the learning rate is multiplied by 0.1 after every 30 epochs.
For \msmt, we divide the learning rate by 10 after every 20 epochs and the total epoch is 50.
Our baseline is based on the Cluster-Contrast~\cite{dai2021cluster}, which achieves great performance.
Differently, 
we utilize the hardest sample in one mini-batch to update the memory bank on \market,
while using all batch samples on \msmt\footnote{This setting is based on our re-implementation of Cluster-Contrast~\cite{dai2021cluster}. 
In fact, we find that the baseline performance drops a lot when directly using the hardest sample for memory update on \msmt.
The setting of the hardest sample for memory updating only achieves 19.5\%/41.2\% mAP/top-1, which largely lower than that of the all batch samples for memory updating (30.1\%/58.6\%). 
}.

\begin{table}
\small
\begin{center}
\setlength{\tabcolsep}{0.8mm}{
\begin{tabu} to 0.9\linewidth {l|X[c]|X[c]|X[c]|X[c]|X[c]|X[c]|X[c]}
\hline
\multicolumn{1}{c|}{\multirow{2}{*}{Method}} & \multirow{2}{*}{No.} & \multicolumn{2}{c|}{\multirow{1}{*}{Components}} & \multicolumn{2}{c|}{\market} & \multicolumn{2}{c}{\msmt} \\
\cline{3-8}
 & & \OurMethodThree & \OurMethodTwo & mAP & top-1 & mAP & top-1 \\ %
\hline
\hline
Baseline & No.1 & - & - & 82.5 & 92.5 & 30.1 & 58.6 \\
\hline
\multirow{4}{*}{\OurWholeMethod} & No.2 & $\lambda$ & - & 83.9 & 93.9 & 33.5 & 63.9 \\
& No.3 & - & \checkmark & 83.6 & 92.7 & 31.4 & 59.9 \\
& No.4 & $\lambda_0$ & \checkmark & 83.3 & 93.4 & 33.8 & \textbf{64.8} \\
& No.5 & $\lambda$ & \checkmark & \textbf{84.7} & \textbf{94.0} & \textbf{35.0} & 64.7 \\
\hline
\end{tabu}}
\end{center}
\vspace{-0.3cm}
\caption{The effectiveness of each component in our implicit sample extension (\OurWholeMethod). 
\OurWholeMethod includes the progressive linear interpolation (\OurMethodThree) and the label-preserving scheme (\OurMethodTwo).
$\lambda_0$ is the constant base degree in \OurMethodThree, and $\lambda$ is the progressively updated one.
}
\label{tab:component_importance}
\end{table}

\begin{figure}[t!]
\centering
\includegraphics[trim =0mm 0mm 0mm 0mm, clip, width=1.0\linewidth]{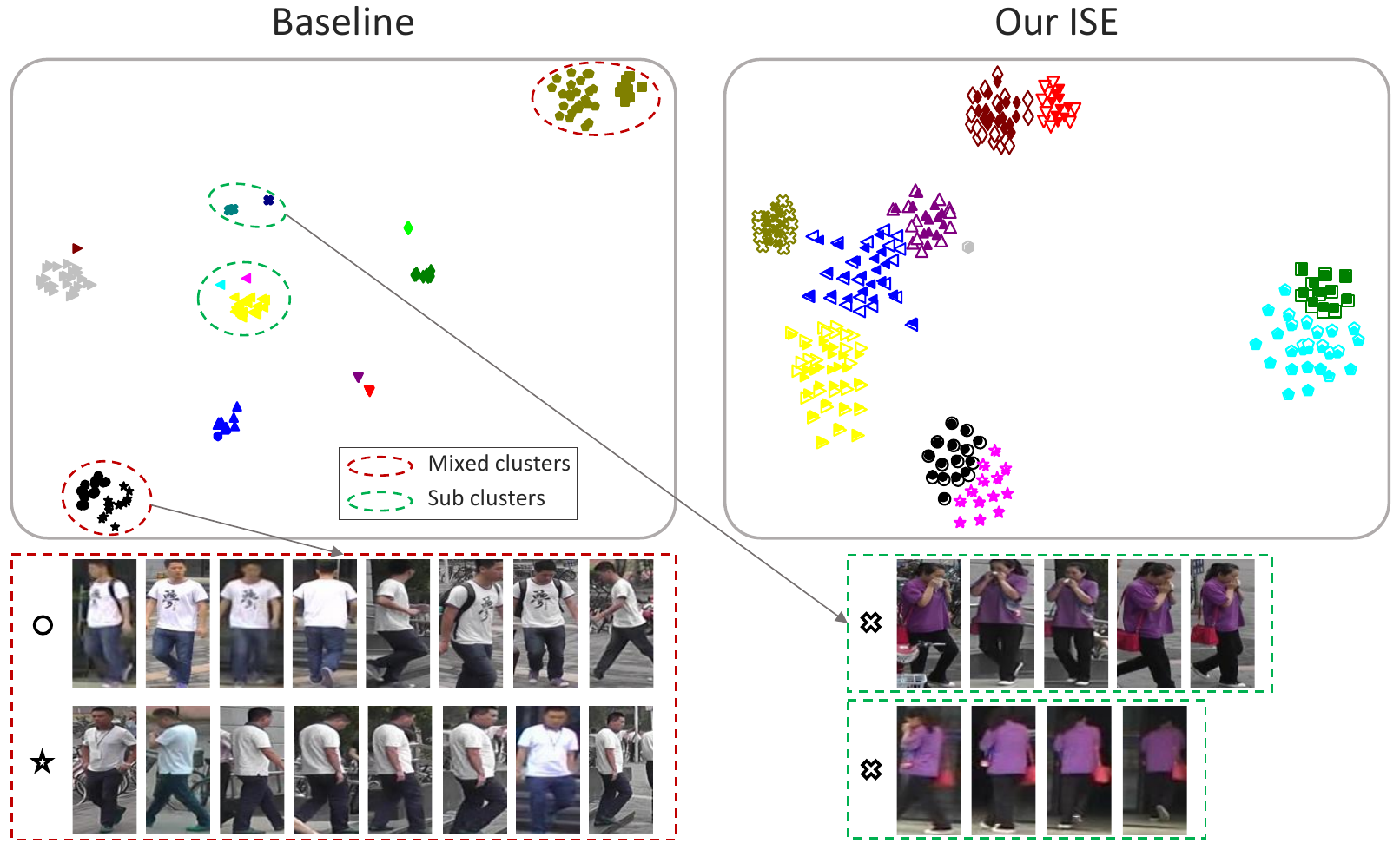}
\caption{
The t-SNE~\cite{van2014accelerating} visualization of clustering results of baseline and our \OurWholeMethod on same samples and epochs.
Except for real samples, we draw the generated support samples (the hollow shapes), to better show the effectiveness of \OurWholeMethod.
Different shapes and colors represent diffident ground-truth identities and pseudo labels, respectively. All inaccurate clusters of these samples are corrected by \OurWholeMethod.
(Best viewed in color and magnification.)
}
\label{fig:ise_behavior}
\end{figure}

\subsection{Ablation Study}
In this subsection, we thoroughly analyze the effectiveness of each component in our \OurWholeMethod, and also 
illustrate the impact of ISE on the clustering quality.

\begin{figure*}[t!]
\centering
\includegraphics[trim =0mm 0mm 0mm 0mm, clip, width=0.9\linewidth]{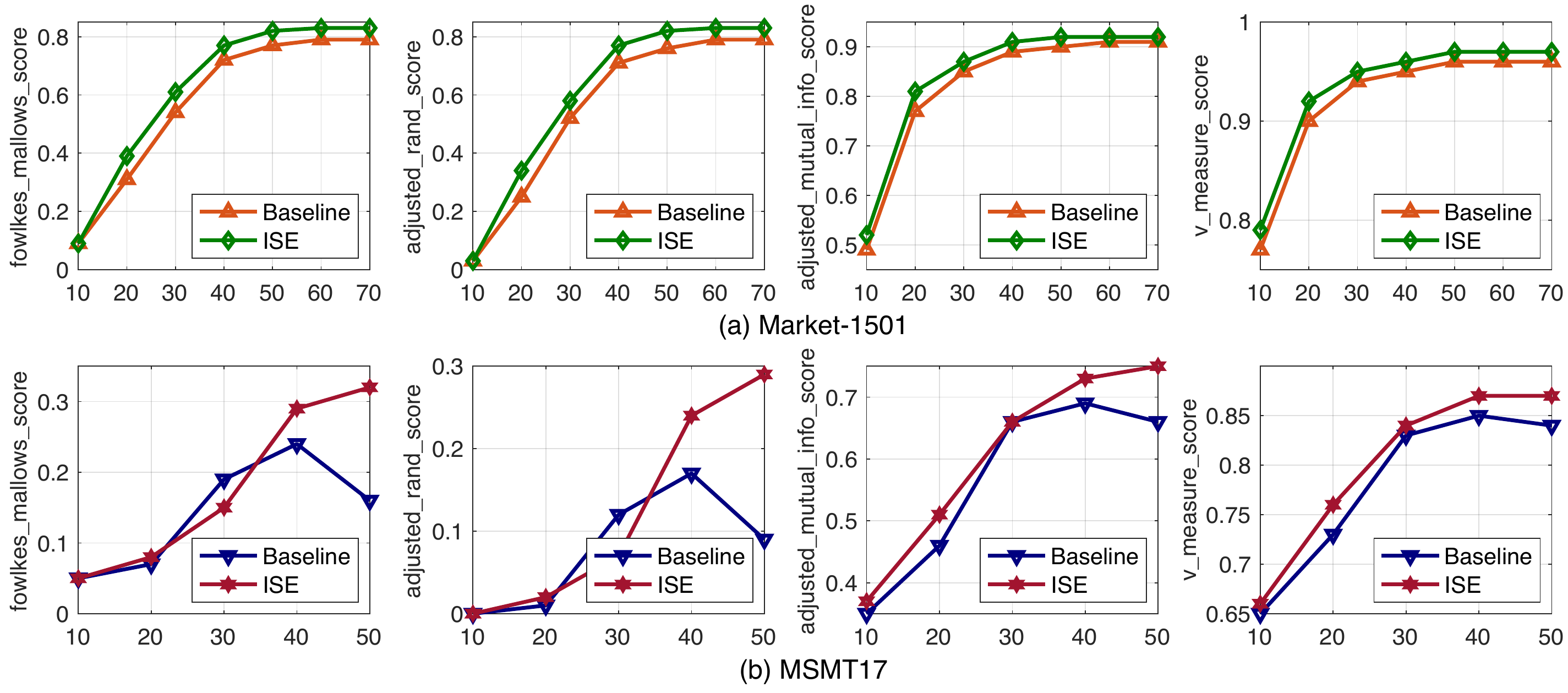}
\setlength{\abovecaptionskip}{-0.02cm} 
\setlength{\belowcaptionskip}{-0.33cm}  %
\caption{
Clustering quality over different epochs from the baseline and our \OurWholeMethod on (a) \market and (b) \msmt datasets.
}
\label{fig:cluster_quality}
\end{figure*}

\myparagraph{Effectiveness of the progressive linear interpolation (\OurMethodThree).}
In Table~\ref{tab:component_importance}, the efficacy of our \OurMethodThree strategy is revealed when comparing (No.1 \vs No.2) and (No.4 \vs No.5).
With the generated support samples from \OurMethodThree, 
the performance largely outperforms the baseline,
especially +1.4\%/+1.4\% mAP/top-1 improvement on \market and
+3.4\%/+5.3\% mAP/top-1 improvement on \msmt.
It clearly shows that our \OurMethodThree is beneficial for improving the feature representation.
We also verify the effectiveness of the progressive update scheme on $\lambda$ (\ie, Eq.~\ref{eq:se_degree}).
When comparing No.5 with No.4, this scheme improves the performance by +1.4\% and +1.2\% mAP on \market and \msmt, showing that gradually farther support samples are necessary to improve the model performance.

\myparagraph{Effectiveness of the label-preserving scheme (LP).}
The results of (No.1 \vs No.3) and (No.2 \vs No.5)
demonstrate the efficacy of \OurMethodTwo.
The No.3 denotes that we apply \OurMethodTwo on actual samples.
In this way, $\widetilde{\pmb{f}}^{+}$ is reduced to $\pmb{f}^{+}$ and $\widetilde{\pmb{f}}^{-}_{c}$ is reduced to $\pmb{f}^{-}_{c}$ in Eq.~\ref{eq:lp_loss}, representing that positive and negative samples are chosen from the actual samples in a mini-batch.
Compared with baseline, the results of No.3 show that \OurMethodTwo leads to +1.1\% and +1.3\% mAP improvement on \market and \msmt, indicating \OurMethodTwo can improve the feature representation even on the actual samples.
When applying \OurMethodTwo on support samples (No.2 \vs No.5), the performance improves from 83.9\% to 84.7\% mAP on \market and 33.5\% to 35.0\% mAP on \msmt.

When combining \OurMethodThree and \OurMethodTwo together,
our \OurWholeMethod achieves +2.2\%/+1.5\% and +4.9\%/+6.1\% mAP/top-1 improvements on \market and \msmt compared to the baseline.
We believe that our \OurWholeMethod indeed generates informative support samples as complements for the data distribution, which is beneficial to improve the model discrimination.

\begin{table}
\small
\begin{center}
\setlength{\tabcolsep}{1.8mm}{
\begin{tabu} to 0.9\linewidth {l|l|c|c|c|c}
\hline
\multirow{2}{*}{Method} & \multirow{2}{*}{Type of $\lambda$} & \multicolumn{2}{c|}{\market} & \multicolumn{2}{c}{\msmt} \\
\cline{3-6}
\cline{3-6}
 & & mAP & top-1 & mAP & top-1 \\ 
\hline
\hline
Baseline & - & 82.5 & 92.5 & 30.1 & 58.6 \\
\hline
\multirow{4}{*}{\OurWholeMethod} & Constant & 83.3 & 93.4 & 33.8 & \textbf{64.8} \\
 & Square & 84.0 & 93.1 & 32.4 & 61.0 \\
 & Linear & 84.6 & 93.7 & 33.8 & 63.0\\
 & Logarithm & \textbf{84.7} & \textbf{94.0} & \textbf{35.0} & 64.7 \\
\hline
\end{tabu}}
\end{center}
\vspace{-0.5cm}
\caption{Comparison with different types of degree $\lambda$ in \OurMethodThree}
\label{tab:lambda_type}
\end{table}

\myparagraph{Clustering quality.}
We intuitively visualize the behavior of support samples.
As shown in Figure~\ref{fig:ise_behavior}, the proposed \OurWholeMethod effectively improves the clustering quality, in which sub clusters and mixed clusters are well alleviated.
We also evaluate the clustering quality over different training epochs on (a) \market and (b) \msmt datasets in Figure~\ref{fig:cluster_quality}.
We employ four metrics from~\cite{clusterquality}\footnote{\href{https://scikit-learn.org/stable/modules/clustering.html\#clustering-performance-evaluation}{{ scikit-learn.clustering-performance-evaluation}}},
including fowlkes\_mallows\_score, adjusted\_rand\_score, adjusted\_mutual\_info\_score and v\_measure\_score.
The larger score represents the better result in all metrics.
It shows that our \OurWholeMethod achieves superior performance than baseline on all metrics. 
In addition, the clustering quality of our \OurWholeMethod progressively improves along the training process, while it may get worse in the baseline.
We believe that the generated support samples can help refine the data distribution in the embedding space, and thus improve the clustering quality.

\begin{table}
\small
\begin{center}
\setlength{\tabcolsep}{1.5mm}{
\begin{tabu} to 0.9\linewidth {l|c|c|c|c|c|c}
\hline
\multirow{2}{*}{Method} & \multirow{2}{*}{Direction} & \multirow{1}{*}{Base}& \multicolumn{2}{c|}{\market} & \multicolumn{2}{c}{\msmt} \\
\cline{4-7}
\cline{4-7}
 & & degree $\lambda_0$ & mAP & top-1 & mAP & top-1 \\ 
\hline
\hline
Baseline & - & - & 82.5 & 92.5 & 30.1 & 58.6 \\
\hline
\multirow{7}{*}{\OurWholeMethod} & Random & \multirow{3}{*}{1.0} & 83.8 & 92.7 & 28.3 & 56.5 \\
 & Farthest &  & 83.8 & 93.1 & 27.8 & 56.2 \\
 & Nearest &  & \textbf{84.7} & \textbf{94.0} & \textbf{35.0} & \textbf{64.7} \\
\cline{2-7}
 & \multirow{4}{*}{Nearest} & 0.1 & 83.2 & 92.4 & 31.2 & 60.0 \\
 &  & 0.5 & 83.6 & 93.1 & 33.0 & 62.3 \\
 &  & 1.0 & \textbf{84.7} & \textbf{94.0} & \textbf{35.0} & \textbf{64.7} \\
 &  & 2.0 & 84.2 & 93.6 & 33.0 & 64.0 \\
\hline
\end{tabu}}
\end{center}
\vspace{-0.5cm}
\caption{Direction and degree of sample extension in \OurWholeMethod}
\label{tab:direction_degree}
\end{table}

\subsection{Parameter Analysis}
\myparagraph{Progressive update type of degree $\lambda$ in \OurMethodThree.}
As shown in Eq.~\ref{eq:se_degree}, we let the degree $\lambda$ increase logarithmically as the training goes on.
In addition to the logarithm type, $\lambda$ can be set as various forms.
We explore four types, \ie, constant ($\lambda=\frac{\lambda_0}{2}$), linear ($\lambda=\frac{\lambda_0\cdot t}{2T}$), square ($\lambda=\frac{\lambda_0 \cdot t^2}{2T^2}$) and logarithm (Eq.~\eqref{eq:se_degree}).
Here we set $\lambda_0=1.0$ for all types.
In Table~\ref{tab:lambda_type}, \OurWholeMethod consistently surpasses the baseline whatever the update type of $\lambda$ is.
By comparison, the performance of progressive types (\eg, linear, logarithm) are usually better than the constant one.
It verifies the assumption that aggressive support sample generation in the early stage may hamper the model optimization.
Table~\ref{tab:lambda_type} also shows that the logarithm type obtains the best results on both \market and \msmt.
Without specification, we use the logarithm as the default update type for $\lambda$ in all other experiments. %

\begin{table}
\small
\begin{center}
\setlength{\tabcolsep}{1.8mm}{
\begin{tabu} to 0.9\linewidth {l|c|c|c|c|c}
\hline
\multirow{2}{*}{Method} & \multirow{1}{*}{$K$-nearest} & \multicolumn{2}{c|}{\market} & \multicolumn{2}{c}{\msmt} \\
\cline{3-6}
 & clusters & mAP & top-1 & mAP & top-1 \\ 
\hline
\hline
Baseline & - & 82.5 & 92.5 & 30.1 & 58.6 \\
\hline
\multirow{4}{*}{\OurWholeMethod} & 1 & 84.7 & \textbf{94.0} & \textbf{35.0} & \textbf{64.7} \\
 & 3 & 84.2 & 93.3 & 33.3 & 63.1 \\
 & 5 & \textbf{84.8} & 93.9 & 31.9 & 61.8 \\
 & 10 & 84.2 & 93.6 & 30.5 & 60.2\\
\hline
\end{tabu}}
\end{center}
\vspace{-0.5cm}
\caption{Influences of the $K$-nearest clusters in our \OurWholeMethod}
\vspace{0.1cm}
\label{tab:topk}
\end{table}

\myparagraph{Direction and degree in \OurMethodThree.}
The generation direction and degree are two elements during the sample extension in our \OurMethodThree.
In this subsection, we first verify the necessity of the $K$-nearest direction, and then explore the influences of different base degree $\lambda_0$. %
For the direction, except for selecting $K$-nearest clusters, we also explore the $K$-random and $K$-farthest selection strategies.
Here we set $K$=1.
The upper block in Table~\ref{tab:direction_degree} shows that the nearest direction obtains the best results.
This is because that support samples distributed between two similar clusters may possibly compensate for some lost yet useful information. By contrast, other two strategies are likely to generate meaningless samples.
Therefore, support samples on various directions can only bring limited improvement and even degradation.
For the degree, we set $\lambda_0=\{0.1, 0.5, 1.0, 2.0\}$ in the lower block in Table~\ref{tab:direction_degree}.
With the increase of $\lambda_0$, the performance gradually improves and then begins to decline. Too large $\lambda_0$ results in farther support samples, which may act as outliers and have negative influences on the model training.  
On the contrary, too small $\lambda_0$ is prone to generating useless support samples, which has less positive impact on the performance.

\begin{table}[t!]
\small
\begin{center}
\setlength{\tabcolsep}{2.2mm}{
\begin{tabu} to 0.9\linewidth {l|c|c|c|c}
\hline
\multirow{2}{*}{Method} & \multicolumn{2}{c|}{MS$\rightarrow$M} & \multicolumn{2}{c}{M$\rightarrow$MS} \\
\cline{2-5}
 & mAP & top-1 & mAP & top-1 \\
\hline
Baseline & 25.5 & 53.8 & 1.5 & 5.9 \\
\OurWholeMethod & \textbf{30.2} & \textbf{59.7} & \textbf{2.7} & \textbf{8.6} \\
\hline
\end{tabu}}
\end{center}
\vspace{-0.5cm}
\caption{Results of the direct domain generalization. ``M'' and ``MS'' represents \market and \msmt, respectively.
}
\vspace{0.1cm}
\label{tab:dg}
\end{table}

\begin{table*}[t!]
\small
\scalebox{0.88}{
\begin{tabular}{l|c|cccc|cccc}
\hline
\multicolumn{1}{l|}{\multirow{2}{*}{Method}} & \multicolumn{1}{c|}{\multirow{2}{*}{Reference}} & \multicolumn{4}{c}{\market} & \multicolumn{4}{|c}{\msmt} \\ 
\cline{3-10}
\multicolumn{1}{c|}{} &\multicolumn{1}{c|}{}& \multicolumn{1}{c}{mAP} & \multicolumn{1}{c}{top-1} & \multicolumn{1}{c}{top-5} & \multicolumn{1}{c|}{top-10} & \multicolumn{1}{c}{mAP} & \multicolumn{1}{c}{top-1}  & \multicolumn{1}{c}{top-5} & \multicolumn{1}{c}{top-10}\\ 
\hline
\hline
\multicolumn{2}{l}{\textit{Purely Unsupervised}}\\
\hline
BUC \cite{lin2019bottom} &AAAI'19&29.6&61.9&73.5&78.2&-&-&-&-\\ 
SSL \cite{lin2020unsupervised}&CVPR'20&37.8&71.7&83.8&87.4&-&-&-&-\\
JVTC \cite{li2020joint}&ECCV'20&41.8&72.9&84.2&88.7&15.1&39.0&50.9&56.8\\
MMCL \cite{wang2020unsupervised} &CVPR'20 &45.5&80.3&89.4&92.3&11.2&35.4&44.8&49.8\\
HCT \cite{zeng2020hierarchical}&CVPR'20 &56.4&80.0&91.6&95.2&-&-&-&-\\
CycAs \cite{wang2020CycAs}&ECCV'20& 64.8&84.8&-&-&26.7&50.1&-&-\\
GCL \cite{chen2021joint}&CVPR'21&66.8&87.3&93.5&95.5&21.3&45.7&58.6&64.5\\
SpCL \cite{ge2020self}&NeurIPS'20 &73.1&88.1&95.1&97.0&19.1&42.3&55.6&61.2\\
IICS~\cite{xuan2021intra} & CVPR'21 & 72.9 & 89.5 & 95.2 & 97.0 & 26.9 & 56.4 & 68.8 & 73.4 \\
JVTC+*~\cite{chen2020joint} & CVPR'21 & 75.4 & 90.5 & 96.2 & 97.1 & 29.7 & 54.4 & 68.2 & 74.2 \\
JNTL-MCSA~\cite{yang2021joint} & CVPR'21 & 61.7 & 83.9 & 92.3 & - & 15.5 & 35.2 & 48.3 & - \\
ICE\cite{chen2021ice}&ICCV'21&79.5&92.0&97.0&98.1&29.8&59.0&71.7&77.0\\
CAP$^{\dagger}$~\cite{wang2020camera}&AAAI'21&79.2&91.4&96.3&97.7&36.9&67.4&78.0&81.4\\
ICE$^{\dagger}$~\cite{chen2021ice} &ICCV'21&\textcolor{black}{82.3}&\textcolor{black}{93.8}&\textcolor{black}{97.6}&\textcolor{black}{98.4}&{38.9}&{70.2}&{80.5}&{84.4}\\
OPLG-HCD~\cite{zheng2021online} & ICCV'21 & 78.1 & 91.1 & 96.4 & 97.7 &26.9 & 53.7 & 65.3 & 70.2 \\
MPRD~\cite{ji2021meta} & ICCV'21 & 51.1 & 83.0 & 91.3 & 93.6 & 14.6 & 37.7 & 51.3 & 57.1 \\
Cluster-Contrast~\cite{dai2021cluster}&Arxiv'21&\underline{82.6}&\underline{93.0}&\underline{97.0}&\underline{98.1}&\underline{33.3}&\underline{63.3}&\underline{73.7}&\underline{77.8}\\ %
\hline
Cluster-Contrast (Our Baseline) & - & 82.5 & 92.5 & 96.9 & 97.9 &  \textcolor{black}{30.1} & \textcolor{black}{58.6} & \textcolor{black}{69.6} & \textcolor{black}{74.4}\\
\textbf{Our \OurWholeMethod}&-&\textbf{84.7}&\textbf{94.0}&\textbf{97.8}&\textbf{98.8}&\textbf{35.0}&\textbf{64.7}&\textbf{75.5}&\textbf{79.4}\\
\textbf{Our \OurWholeMethod + GeM pooling} &-&\underline{\textbf{85.3}}&\underline{\textbf{94.3}}&\underline{\textbf{98.0}}&\underline{\textbf{98.8}}&\underline{\textbf{37.0}}&\underline{\textbf{67.6}}&\underline{\textbf{77.5}}&\underline{\textbf{81.0}}\\
\hline
\multicolumn{2}{l}{\textit{Fully Supervised}}\\
\hline
PCB \cite{sun2018beyond}&ECCV'18&81.6&93.8&97.5&98.5&40.4&68.2&-&-\\
DG-Net \cite{zheng2019joint}&CVPR'19&86.0&94.8&-&-&52.3&77.2&-&-\\
ICE (w/ ground-truth) \cite{chen2021ice}&ICCV'21&86.6&95.1&98.3&98.9&50.4&76.4&86.6&90.0\\
Our \OurWholeMethod (w/ ground-truth) &-&87.8&95.6&98.5&99.2&51.0&76.8&87.1&90.6\\
\hline
\end{tabular}}
\centering
\vspace{-0.2cm}
\caption{Comparison of ReID methods on \market and \msmt datasets.
The best USL results without camera information are marked with \textbf{bold}.
The pure \textbf{bold} number represents using the average pooling, while the \underline{underline} number denotes using the GeM~\cite{radenovic2018fine} pooling.
$\dagger$ indicates using the additional camera knowledge.
}
\vspace{-0.5cm}
\label{tab:sota}
\end{table*}

\myparagraph{Sample extension to $K$-nearest clusters.}
Table~\ref{tab:topk} shows the influences of different number of $K$ in our method.
We set $K=\{1, 3, 5, 10\}$ in this experiment.
We can see that as $K$ grows, the performance almost remains the same on \market but drops a lot on \msmt.
Meanwhile, too large $K$ is also harmful to the performance on \market.
We conjecture it may be because the data points are distributed in a manifold, which requires linear interpolation to be within the local Euclidean space near each point. 
If $K$ is large, the $K$-th centroid is far from the point-of-interest
and even out of the local Euclidean space near the given point.
The generated support sample is thus not in the original manifold and does not belong to any cluster.
Consequentially, it involves noises and decreases the performance.
Thus, we set $K=1$ in this paper.

\begin{figure}[t]
\centering
\includegraphics[trim =0mm 0mm 0mm 0mm, clip, width=1.0\linewidth]{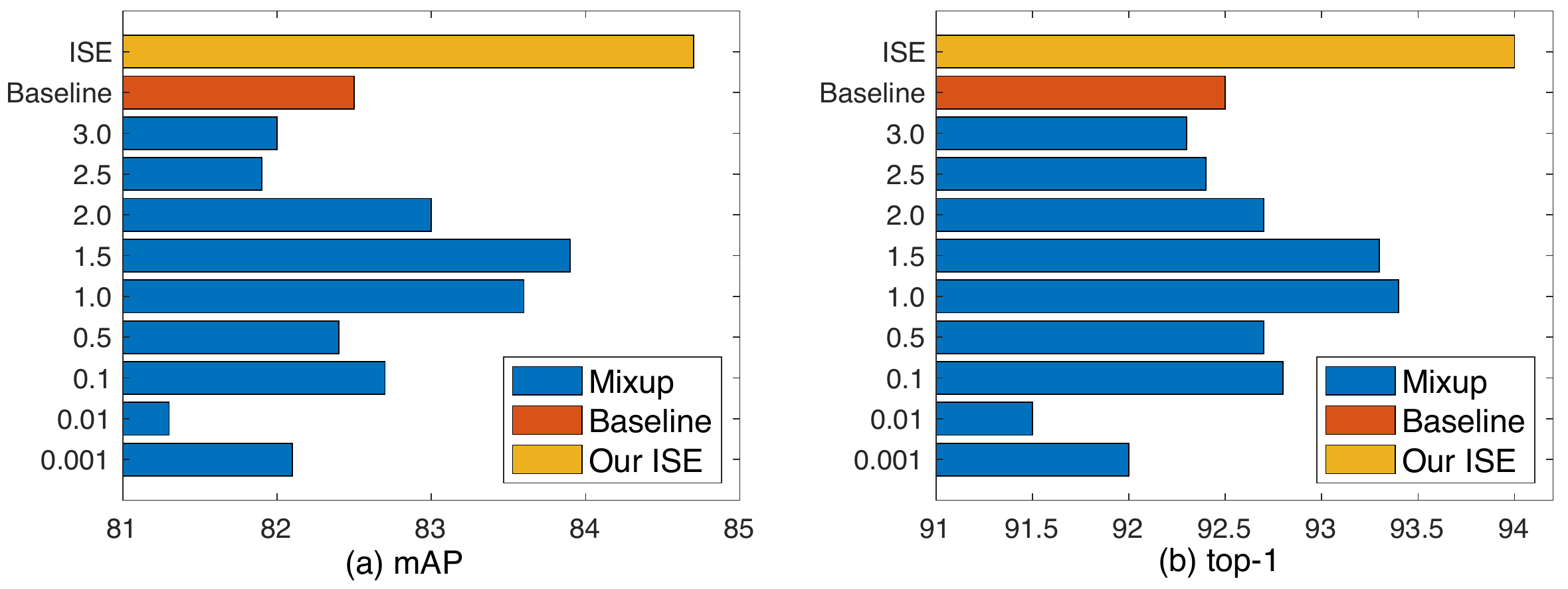}
\setlength{\abovecaptionskip}{-0.1cm} 
\caption{
Our \OurWholeMethod \vs traditional Mixup~\cite{zhang2017mixup} on \market.
The number of y-axis is the hyper-parameter $\alpha$, representing the interpolation ratio randomly sampled from $\lambda\sim \mathrm{Beta}(\alpha ,\alpha )$.
}
\label{fig:mixup}
\end{figure}

\subsection{More Discussions}
\myparagraph{Our \OurWholeMethod \vs Mixup~\cite{zhang2017mixup}.}
Although ISE and Mixup both generate interpolated samples, the differences are significant from the following three aspects: 
1) The interpolation of Mixup takes place on the raw images or hidden features, while ours is on the output embeddings. 
2) The interpolation in Mixup can be between two arbitrary inputs, while ISE chooses the $K$-nearest centroids as the interpolating direction. 
3) The interpolated samples in Mixup are with interpolated labels and do not belong to any existing classes. In contrast, support samples in ISE belong to a single identified class controlled by the generation degree.
In Figure~\ref{fig:mixup}, we experimentally investigate Mixup and compare it with our \OurWholeMethod.
With an appropriate hyper-parameter $\alpha$, Mixup can improve the performance due to its data augmentation effect on input images.
However, the improvement is limited without the controlment on the direction and degree of synthetics. 
Differently, our \OurWholeMethod can effectively generate informative support samples between two similar clusters in the embedding space to compensate for missing information.

\myparagraph{\OurWholeMethod helps domain generalization (DG).}
We find that several clusters in baseline easily shrink to a small region in the embedding space as shown in Figure~\ref{fig:ise_behavior}. Such compact data distribution increases the risk of overfitting. By contrast, our \OurWholeMethod generates support samples that patrol around cluster boundaries and improve clustering behaviours. We think this should help domain generalization as well. Thus we investigate how the baseline and our \OurWholeMethod perform in DG tasks.
As shown in Table~\ref{tab:dg}, our \OurWholeMethod significantly outperforms the baseline and obtains consistent superior results.
Especially, \OurWholeMethod brings +4.7\%/+5.9\% mAP/top-1 improvements on the MS$\rightarrow$M setting.
This demonstrates that support samples indeed help refine the data distribution and improve the generalization ability of the Re-ID model.

\subsection{Comparison with State-of-the-art Methods}
We compare the proposed \OurWholeMethod with state-of-the-art person Re-ID methods in Table~\ref{tab:sota}.
For the USL setting,
\OurWholeMethod is superior or competitive to the previous methods.
In particular, our method significantly outperforms other previous methods with average pooling.
\OurWholeMethod obtains 84.7\% mAP and 94.0\% top-1 on \market dataset and 35.0\% mAP and 64.7\% top-1 on \msmt dataset.
Considering that Cluster-Contrast~\cite{dai2021cluster} utilizes the generalized mean pooling (GeM)~\cite{radenovic2018fine} in the final results\footnote{It is noticing that we carefully check the code and the released training logs of Cluster-Contrast, and find they use GeM pooling in the final results.
For a fair comparison, we also report the results of \OurWholeMethod + GeM pooling.
Here, we report the results of the first version of Cluster-Contrast~\cite{dai2021cluster}.
}, we also report \OurWholeMethod's results with GeM for a fair comparison.
With GeM, our \OurWholeMethod still consistently achieves better results.
Note that different from ICE~\cite{chen2021ice} and CAP~\cite{wang2020camera},
we do not use the camera information.
Under the non-camera setting, our \OurWholeMethod obtains 35.0\%/64.7\% mAP/top-1 on \msmt, largely outperforming OPLG-HCD~\cite{zheng2021online}, ICE~\cite{chen2021ice} (without cameras) and Cluster-Contrast~\cite{dai2021cluster}.
In addition, we also transfer our \OurWholeMethod to the supervised setting by using ground-truth identities instead of pseudo labels.
From the lower block in Table~\ref{tab:sota}, we can see that our \OurWholeMethod is competitive with well-known supervised methods (\eg, PCB and DG-Net) and surpasses ICE~\cite{chen2021ice} with ground-truths.
The results show the effectiveness of \OurWholeMethod on supervised Re-ID.

\section{Conclusion and Limitations}
\myparagraph{Conclusion.}
In the clustering-based unsupervised person Re-ID methods,
when one identity breaks into multiple clusters or different identities get mixed into one cluster, they will confuse the learning and result in unsatisfying performance. To alleviate this issue, we propose an implicit sample extension method to generate support samples between two neighbouring clusters, and expect to associate the same identity's samples by these support samples. Specifically, we develop a progressive linear interpolation strategy to control the generation of support sample with direction and degree factors. A label-preserving loss is also proposed to constrain support samples close to their own cluster members. Comprehensive experiments show that our method can improve clustering results significantly and achieve new state-of-the-art performance on two popular Re-ID datasets. The proposed method also demonstrates better domain generalization ability than the baseline. 

\myparagraph{Limitations.}
Our \OurWholeMethod generates support samples from actual samples to their neighboring cluster centroids in the embedding space.
For each sample, the number of the selected nearest cluster centroids is fixed to $K$.
It may not be the optimal solution.
If some clusters have been well classified, they do not need to add supplementary samples.
In contrast, when a specific cluster is mixed by multiple identities, it is better to generate support samples in multiple $K$-nearest directions.
We will explore a dynamic $K$ in the future work.

\myparagraph{Acknowledgements} This work was supported in part by 
the Major Project for New Generation of AI (No.2018AAA0100400),
the National Natural Science Foundation of China (No. 61836014, No. U21B2042, No. 62072457, No. 62006231).

\clearpage
{\small
\bibliographystyle{ieee_fullname}
\bibliography{egbib}

\begin{thebibliography}{10}\itemsep=-1pt

\bibitem{bai2017scalable}
Song Bai, Xiang Bai, and Qi Tian.
\newblock Scalable person re-identification on supervised smoothed manifold.
\newblock In {\em Proc. IEEE Conf. Comp. Vis. Patt. Recogn.}, 2017.

\bibitem{UMSDA}
Zechen Bai, Zhigang Wang, Jian Wang, Di Hu, and Errui Ding.
\newblock Unsupervised multi-source domain adaptation for person
  re-identification.
\newblock In {\em Proc. IEEE Conf. Comp. Vis. Patt. Recogn.}, page
  12914–12923. Computer Vision Foundation / {IEEE}, 2021.

\bibitem{bengio2013better}
Yoshua Bengio, Gr{\'e}goire Mesnil, Yann Dauphin, and Salah Rifai.
\newblock Better mixing via deep representations.
\newblock In {\em Proc. Int. Conf. Mach. Learn.}, pages 552--560. PMLR, 2013.

\bibitem{chen2021ice}
Hao Chen, Benoit Lagadec, and Francois Bremond.
\newblock Ice: Inter-instance contrastive encoding for unsupervised person
  re-identification.
\newblock {\em Proc. IEEE Int. Conf. Comp. Vis.}, 2021.

\bibitem{chen2020joint}
Hao Chen, Yaohui Wang, Benoit Lagadec, Antitza Dantcheva, and Francois Bremond.
\newblock Joint generative and contrastive learning for unsupervised person
  re-identification.
\newblock {\em Proc. IEEE Conf. Comp. Vis. Patt. Recogn.}, 2021.

\bibitem{chen2021joint}
Hao Chen, Yaohui Wang, Benoit Lagadec, Antitza Dantcheva, and Francois Bremond.
\newblock Joint generative and contrastive learning for unsupervised person
  re-identification.
\newblock In {\em Proc. IEEE Conf. Comp. Vis. Patt. Recogn.}, pages 2004--2013,
  2021.

\bibitem{clusterquality}
David Cournapeau and Google members.
\newblock scikit-learn.
\newblock \url{https://scikit-learn.org/stable/index.html}, 2007.

\bibitem{dai2021idm}
Yongxing Dai, Jun Liu, Yifan Sun, Zekun Tong, Chi Zhang, and Ling-Yu Duan.
\newblock Idm: An intermediate domain module for domain adaptive person re-id.
\newblock In {\em Proc. IEEE Int. Conf. Comp. Vis.}, pages 11864--11874, 2021.

\bibitem{dai2021cluster}
Zuozhuo Dai, Guangyuan Wang, Weihao Yuan, Siyu Zhu, and Ping Tan.
\newblock Cluster contrast for unsupervised person re-identification.
\newblock {\em arXiv preprint arXiv:2103.11568}, 2021.

\bibitem{deng2009imagenet}
Jia Deng, Wei Dong, Richard Socher, Li-Jia Li, Kai Li, and Li Fei-Fei.
\newblock Imagenet: A large-scale hierarchical image database.
\newblock In {\em Proc. IEEE Conf. Comp. Vis. Patt. Recogn.}, pages 248--255.
  Ieee, 2009.

\bibitem{deng2018image}
Weijian Deng, Liang Zheng, Qixiang Ye, Guoliang Kang, Yi Yang, and Jianbin
  Jiao.
\newblock Image-image domain adaptation with preserved self-similarity and
  domain-dissimilarity for person re-identification.
\newblock In {\em Proc. IEEE Conf. Comp. Vis. Patt. Recogn.}, pages 994--1003,
  2018.

\bibitem{ester1996density}
Martin Ester, Hans-Peter Kriegel, J{\"o}rg Sander, Xiaowei Xu, et~al.
\newblock A density-based algorithm for discovering clusters in large spatial
  databases with noise.
\newblock In {\em Kdd}, volume~96, pages 226--231, 1996.

\bibitem{fu2019self}
Yang Fu, Yunchao Wei, Guanshuo Wang, Yuqian Zhou, Honghui Shi, and Thomas~S
  Huang.
\newblock Self-similarity grouping: A simple unsupervised cross domain
  adaptation approach for person re-identification.
\newblock In {\em Proc. IEEE Int. Conf. Comp. Vis.}, pages 6112--6121, 2019.

\bibitem{ge2020self}
Yixiao Ge, Feng Zhu, Dapeng Chen, Rui Zhao, and hongsheng Li.
\newblock Self-paced contrastive learning with hybrid memory for domain
  adaptive object re-id.
\newblock In {\em Proc. Advances in Neural Inf. Process. Syst.}, volume~33,
  pages 11309--11321. Curran Associates, Inc., 2020.

\bibitem{gray2007evaluating}
Douglas Gray, Shane Brennan, and Hai Tao.
\newblock Evaluating appearance models for recognition, reacquisition, and
  tracking.
\newblock In {\em Proc. IEEE Int. Workshop on Performance Evaluation for
  Tracking and Surveillance}, volume~3, pages 1--7. Citeseer, 2007.

\bibitem{he2016deep}
Kaiming He, Xiangyu Zhang, Shaoqing Ren, and Jian Sun.
\newblock Deep residual learning for image recognition.
\newblock In {\em Proc. IEEE Conf. Comp. Vis. Patt. Recogn.}, pages 770--778,
  2016.

\bibitem{ji2021meta}
Haoxuanye Ji, Le Wang, Sanping Zhou, Wei Tang, Nanning Zheng, and Gang Hua.
\newblock Meta pairwise relationship distillation for unsupervised person
  re-identification.
\newblock In {\em Proc. IEEE Int. Conf. Comp. Vis.}, pages 3661--3670, 2021.

\bibitem{kingma2014adam}
Diederik~P Kingma and Jimmy Ba.
\newblock Adam: A method for stochastic optimization.
\newblock {\em Int. Conf. Learn. Represent.}, 2014.

\bibitem{li2020joint}
Jianing Li and Shiliang Zhang.
\newblock Joint visual and temporal consistency for unsupervised domain
  adaptive person re-identification.
\newblock In {\em Proc. Eur. Conf. Comp. Vis.}, pages 483--499. Springer, 2020.

\bibitem{lin2019bottom}
Yutian Lin, Xuanyi Dong, Liang Zheng, Yan Yan, and Yi Yang.
\newblock A bottom-up clustering approach to unsupervised person
  re-identification.
\newblock In {\em Proc. AAAI Conf. Artificial Intell.}, volume~33, pages
  8738--8745, 2019.

\bibitem{lin2020unsupervised}
Yutian Lin, Lingxi Xie, Yu Wu, Chenggang Yan, and Qi Tian.
\newblock Unsupervised person re-identification via softened similarity
  learning.
\newblock In {\em Proc. IEEE Conf. Comp. Vis. Patt. Recogn.}, pages 3390--3399,
  2020.

\bibitem{macqueen1967some}
James MacQueen et~al.
\newblock Some methods for classification and analysis of multivariate
  observations.
\newblock In {\em Proceedings of the fifth Berkeley symposium on mathematical
  statistics and probability}, volume~1, pages 281--297. Oakland, CA, USA,
  1967.

\bibitem{oord2018representation}
Aaron van~den Oord, Yazhe Li, and Oriol Vinyals.
\newblock Representation learning with contrastive predictive coding.
\newblock {\em arXiv preprint arXiv:1807.03748}, 2018.

\bibitem{radenovic2018fine}
Filip Radenovi{\'c}, Giorgos Tolias, and Ond{\v{r}}ej Chum.
\newblock Fine-tuning cnn image retrieval with no human annotation.
\newblock {\em {IEEE} Trans. Pattern Anal. Mach. Intell.}, 41(7):1655--1668,
  2018.

\bibitem{udatp}
Liangchen Song, Cheng Wang, Lefei Zhang, Bo Du, Qian Zhang, Chang Huang, and
  Xinggang Wang.
\newblock Unsupervised domain adaptive re-identification: Theory and practice.
\newblock {\em Pattern Recogn.}, 102:107173, 2020.

\bibitem{sun2018beyond}
Yifan Sun, Liang Zheng, Yi Yang, Qi Tian, and Shengjin Wang.
\newblock Beyond part models: Person retrieval with refined part pooling (and a
  strong convolutional baseline).
\newblock In {\em Proc. Eur. Conf. Comp. Vis.}, 2018.

\bibitem{viewmaker}
Alex Tamkin, Mike Wu, and Noah~D. Goodman.
\newblock Viewmaker networks: Learning views for unsupervised representation
  learning.
\newblock In {\em Proc. Int. Conf. Learn. Representations}, 2021.

\bibitem{upchurch2017deep}
Paul Upchurch, Jacob Gardner, Geoff Pleiss, Robert Pless, Noah Snavely, Kavita
  Bala, and Kilian Weinberger.
\newblock Deep feature interpolation for image content changes.
\newblock In {\em Proc. IEEE Conf. Comp. Vis. Patt. Recogn.}, pages 7064--7073,
  2017.

\bibitem{van2014accelerating}
Laurens Van Der~Maaten.
\newblock Accelerating t-sne using tree-based algorithms.
\newblock {\em J. Mach. Learn. Res.}, 15(1):3221--3245, 2014.

\bibitem{wang2020unsupervised}
Dongkai Wang and Shiliang Zhang.
\newblock Unsupervised person re-identification via multi-label classification.
\newblock In {\em Proc. IEEE Conf. Comp. Vis. Patt. Recogn.}, pages
  10981--10990, 2020.

\bibitem{wang2020camera}
Menglin Wang, Baisheng Lai, Jianqiang Huang, Xiaojin Gong, and Xian-Sheng Hua.
\newblock Camera-aware proxies for unsupervised person re-identification.
\newblock {\em Proc. AAAI Conf. Artificial Intell.}, 2021.

\bibitem{wang2021regularizing}
Yulin Wang, Gao Huang, Shiji Song, Xuran Pan, Yitong Xia, and Cheng Wu.
\newblock Regularizing deep networks with semantic data augmentation.
\newblock {\em {IEEE} Trans. Pattern Anal. Mach. Intell.}, 2021.

\bibitem{CysAs}
Zhongdao Wang, Jingwei Zhang, Liang Zheng, Yixuan Liu, Yifan Sun, Yali Li, and
  Shengjin Wang.
\newblock Cycas: Self-supervised cycle association for learning re-identifiable
  descriptions.
\newblock In {\em Proc. Eur. Conf. Comp. Vis.}, volume 12356, pages 72--88,
  2020.

\bibitem{wang2020CycAs}
Zhongdao Wang, Jingwei Zhang, Liang Zheng, Yixuan Liu, Yifan Sun, Yali Li, and
  Shengjin Wang.
\newblock Cycas: Self-supervised cycle association for learning re-identifiable
  descriptions.
\newblock {\em Proc. Eur. Conf. Comp. Vis.}, 2020.

\bibitem{wei2018person}
Longhui Wei, Shiliang Zhang, Wen Gao, and Qi Tian.
\newblock Person transfer gan to bridge domain gap for person
  re-identification.
\newblock In {\em Proc. IEEE Conf. Comp. Vis. Patt. Recogn.}, pages 79--88,
  2018.

\bibitem{TSSL}
Guile Wu, Xiatian Zhu, and Shaogang Gong.
\newblock Tracklet self-supervised learning for unsupervised person
  re-identification.
\newblock In {\em Proc. AAAI Conf. Artificial Intell.}, pages 12362--12369,
  2020.

\bibitem{xuan2021intra}
Shiyu Xuan and Shiliang Zhang.
\newblock Intra-inter camera similarity for unsupervised person
  re-identification.
\newblock In {\em Proc. IEEE Conf. Comp. Vis. Patt. Recogn.}, pages
  11926--11935, 2021.

\bibitem{asymmetric}
Fengxiang Yang, Ke Li, Zhun Zhong, Zhiming Luo, Xing Sun, Hao Cheng, Xiaowei
  Guo, Feiyue Huang, Rongrong Ji, and Shaozi Li.
\newblock Asymmetric co-teaching for unsupervised cross-domain person
  re-identification.
\newblock In {\em Proc. AAAI Conf. Artificial Intell.}, pages 12597--12604,
  2020.

\bibitem{yang2021joint}
Fengxiang Yang, Zhun Zhong, Zhiming Luo, Yuanzheng Cai, Yaojin Lin, Shaozi Li,
  and Nicu Sebe.
\newblock Joint noise-tolerant learning and meta camera shift adaptation for
  unsupervised person re-identification.
\newblock In {\em Proc. IEEE Conf. Comp. Vis. Patt. Recogn.}, pages 4855--4864,
  2021.

\bibitem{cutmix}
Sangdoo Yun, Dongyoon Han, Sanghyuk Chun, Seong~Joon Oh, Youngjoon Yoo, and
  Junsuk Choe.
\newblock Cutmix: Regularization strategy to train strong classifiers with
  localizable features.
\newblock In {\em Proc. IEEE Int. Conf. Comp. Vis.}, pages 6022--6031.

\bibitem{zeng2020hierarchical}
Kaiwei Zeng, Munan Ning, Yaohua Wang, and Yang Guo.
\newblock Hierarchical clustering with hard-batch triplet loss for person
  re-identification.
\newblock In {\em Proc. IEEE Conf. Comp. Vis. Patt. Recogn.}, pages
  13657--13665, 2020.

\bibitem{zhang2017mixup}
Hongyi Zhang, Moustapha Cisse, Yann~N Dauphin, and David Lopez-Paz.
\newblock mixup: Beyond empirical risk minimization.
\newblock In {\em Int. Conf. Learn. Represent.}, 2017.

\bibitem{zhang2019self}
Xinyu Zhang, Jiewei Cao, Chunhua Shen, and Mingyu You.
\newblock Self-training with progressive augmentation for unsupervised
  cross-domain person re-identification.
\newblock In {\em Proc. IEEE Int. Conf. Comp. Vis.}, pages 8222--8231, 2019.

\bibitem{zhang2020memorizing}
Xinyu Zhang, Dong Gong, Jiewei Cao, and Chunhua Shen.
\newblock Memorizing comprehensively to learn adaptively: Unsupervised
  cross-domain person re-id with multi-level memory.
\newblock {\em arXiv preprint arXiv:2001.04123}, 2020.

\bibitem{zheng2015scalable}
Liang Zheng, Liyue Shen, Lu Tian, Shengjin Wang, Jingdong Wang, and Qi Tian.
\newblock Scalable person re-identification: A benchmark.
\newblock In {\em Proc. IEEE Int. Conf. Comp. Vis.}, pages 1116--1124, 2015.

\bibitem{zheng2021online}
Yi Zheng, Shixiang Tang, Guolong Teng, Yixiao Ge, Kaijian Liu, Jing Qin,
  Donglian Qi, and Dapeng Chen.
\newblock Online pseudo label generation by hierarchical cluster dynamics for
  adaptive person re-identification.
\newblock In {\em Proc. IEEE Int. Conf. Comp. Vis.}, pages 8371--8381, 2021.

\bibitem{zheng2019joint}
Zhedong Zheng, Xiaodong Yang, Zhiding Yu, Liang Zheng, Yi Yang, and Jan Kautz.
\newblock Joint discriminative and generative learning for person
  re-identification.
\newblock In {\em Proc. IEEE Conf. Comp. Vis. Patt. Recogn.}, 2019.

\bibitem{zhong2017re}
Zhun Zhong, Liang Zheng, Donglin Cao, and Shaozi Li.
\newblock Re-ranking person re-identification with k-reciprocal encoding.
\newblock In {\em Proc. IEEE Conf. Comp. Vis. Patt. Recogn.}, pages 1318--1327,
  2017.

\bibitem{zhong2019invariance}
Zhun Zhong, Liang Zheng, Zhiming Luo, Shaozi Li, and Yi Yang.
\newblock Invariance matters: Exemplar memory for domain adaptive person
  re-identification.
\newblock In {\em Proc. IEEE Conf. Comp. Vis. Patt. Recogn.}, pages 598--607,
  2019.

\bibitem{zhong2020learning}
Zhun Zhong, Liang Zheng, Zhiming Luo, Shaozi Li, and Yi Yang.
\newblock Learning to adapt invariance in memory for person re-identification.
\newblock {\em IEEE Trans. Pattern Anal. Mach. Intell.}, 2020.

\bibitem{zou2020joint}
Yang Zou, Xiaodong Yang, Zhiding Yu, BVK Kumar, and Jan Kautz.
\newblock Joint disentangling and adaptation for cross-domain person
  re-identification.
\newblock {\em arXiv preprint arXiv:2007.10315}, 2020.

\end{thebibliography}
}

\end{document}